\documentclass[a4paper,twoside]{article}
\usepackage[T1]{fontenc}
\usepackage[utf8]{inputenc}
\usepackage[english]{babel}
\usepackage{graphicx,amsmath,amssymb,booktabs,array,multirow,xurl}
\usepackage{pslatex,apalike,SCITEPRESS}
\usepackage{etoolbox,placeins,microtype,flushend,ragged2e,dblfloatfix,cuted}
\makeatletter
\renewcommand{\title}[1]{\gdef\@title{#1}}
\patchcmd{\maketitle}{\vskip -0.07in}{\vskip 12pt}{}{\PackageError{manuscript}{Title-author spacing patch failed}{Check the SCITEPRESS title definition.}}
\patchcmd{\maketitle}{\vskip 0.51in}{\vskip 15pt}{}{\PackageError{manuscript}{Title spacing patch failed}{Check the SCITEPRESS title definition.}}
\makeatother
\usepackage[hidelinks]{hyperref}
\hypersetup{pdftitle={Initialization and Stopping Tolerance in CPU Dermoscopic Segmentation},pdfauthor={Wenhao Xu, Yixian Kong, Ting Pan, Feilong Wang, Changwei Wang},pdfsubject={Research manuscript}}
\graphicspath{{figures/}}
\newcommand{\DiceCheckDefault}{0.6011}
\newcommand{\DiceSeedDefault}{0.6660}
\newcommand{\DeltaPrimary}{0.0649}
\newcommand{\PrimaryCI}{[0.0452, 0.0860]}
\newcommand{\DiceOtsu}{0.6897}
\newcommand{\DiskOneCount}{471}

\newcommand{\epsd}{10^{-3}}
\newcommand{\epst}{10^{-5}}
\begin{document}
\title{Initialization and Stopping Tolerance in CPU Dermoscopic Segmentation}
\author{\authorname{Wenhao Xu$^{1}$, Yixian Kong$^{2}$, Ting Pan$^{2}$, Feilong Wang$^{2}$, Changwei Wang$^{3,*}$}
\affiliation{$^{1}$Zhengzhou Police University, Zhengzhou, China}
\affiliation{$^{2}$School of Artificial Intelligence, Beijing University of Posts and Telecommunications, Beijing, China}
\affiliation{$^{3}$Qilu University of Technology and Jinan Supercomputing Center, Jinan, China}
\email{*Corresponding author: changweiwang@sdas.org}}
\keywords{Skin Lesion Segmentation, Chan--Vese, Initialization, Stopping Tolerance, Reproducibility.}
\abstract{Contour initialization and numerical stopping can jointly affect the evaluation of active-contour segmentation. We examine their interaction using the open-source scikit-image Chan--Vese implementation on a resized ISIC 2017 mirror. A fixed development set of 100 images selects a common input channel; all 600 images in the repository's held-out partition are then evaluated. Otsu thresholding is compared with checkerboard-, disk-, and Otsu-initialized contours under default and tighter level-set tolerances. At the default tolerance, Otsu initialization increases mean image Dice from \DiceCheckDefault{} to \DiceSeedDefault{} relative to checkerboard initialization, a paired difference of \DeltaPrimary{} (95\% image-bootstrap interval \PrimaryCI{}). Otsu thresholding alone achieves \DiceOtsu{}. The default disk initializer stops after one iteration on \DiskOneCount{} images. Tightening the tolerance reduces the Otsu-seed advantage over checkerboard initialization to 0.0197, with most runs reaching the 500-iteration limit. The default-tolerance advantage also reverses between small- and large-lesion strata. These findings show that an improvement over a generic initializer can coexist with deterioration relative to the threshold baseline. Evaluations should retain the unrefined mask as a comparator and report the initial-field definition, stopping tolerance, and observed iteration counts together.}
\onecolumn\maketitle\normalsize
\raggedbottom

\section{INTRODUCTION}
\label{sec:intro}
Dermoscopic lesion segmentation delineates a lesion against surrounding skin and forms a distinct task in the ISIC challenge \cite{codella2017}. Learned approaches can optimize overlap directly through a Jaccard-based objective \cite{yuan2017}; generalized Dice losses address foreground imbalance in medical segmentation \cite{sudre2017}. Training-free methods remain useful reference points, particularly when computation is limited to a CPU. A useful comparison should measure the contribution of each processing step.

Otsu thresholding estimates a global intensity split \cite{otsu1979}, whereas the Chan--Vese model evolves a contour through regional intensity agreement and a boundary penalty \cite{chan2001}. Using a threshold mask to initialize the contour connects the two approaches. The practical question is whether contour evolution improves this partition enough to justify its additional computation.

The answer depends on the numerical implementation as well as the energy model. A level-set solver requires an initial field, update rule, stopping tolerance, and iteration limit. A tolerance on changes in the field does not directly measure segmentation error. Different initial fields can therefore receive substantially different amounts of contour evolution under the same stopping rule.

This study evaluates three Chan--Vese initializers and an Otsu-only baseline on 600 held-out image IDs. A development-motivated sensitivity analysis changes only the stopping tolerance and records the resulting iteration counts. The comparison links overlap, boundary agreement, computation, and lesion size under a common pipeline. Source snapshots, data hashes, predictions, and per-image measurements accompany the manuscript.

\section{RELATED WORK}
\label{sec:related}
\subsection{Active Contours and Initialization}
Classical snakes combine curve regularity with image forces \cite{kass1988}; geodesic active contours express boundary detection through geometric curve evolution \cite{caselles1997}. Chan--Vese instead fits regional intensity means, allowing segmentation without a strong image gradient \cite{chan2001}. Localized region models and region-scalable fitting relax the global intensity assumption \cite{lankton2008,li2008rsf}. They address spatial inhomogeneity by changing the fitting model, so their behavior cannot be inferred by tightening a global model's stopping tolerance.

Numerical treatment is a separate consideration. Distance-regularized evolution controls the level-set profile during optimization \cite{li2010drlse}, while Getreuer provides a reproducible treatment of Chan--Vese segmentation \cite{getreuer2012}. We use the public scikit-image solver \cite{walt2014,skimagecode} and vary its initialization and stopping tolerance without modifying the energy or update rule.

For dermoscopy, previous studies compare contour initializers \cite{nagieb2018}, optimize the initial circular region with a genetic algorithm \cite{ashour2021}, and use adaptive Otsu-based initialization \cite{malik2022}. The present experiment evaluates a plain histogram threshold and its signed-distance field; it does not reproduce the adaptive algorithm. Retaining the threshold mask as a separate comparator isolates the contribution of contour evolution.

\subsection{Learned Context and Boundaries}
U-Net couples an encoder--decoder with skip connections \cite{ronneberger2015}; UNet++, Attention U-Net, and DoubleU-Net extend feature transfer, gating, and network staging \cite{zhou2018,oktay2018,jha2020}. Recurrent residual units accumulate features in R2U-Net \cite{alom2018r2unet}, while MultiResUNet combines multiple receptive-field scales and residual paths \cite{ibtehaz2019multires}. CE-Net uses atrous convolution and multi-kernel pooling to retain spatial context \cite{gu2019cenet}, and \mbox{UNet~3+} aggregates full-scale features with deep supervision \cite{huang2020unet3plus}. These designs address spatial information that a two-mean intensity model does not represent.

Transformer approaches include hybrid convolution--attention models \cite{chen2021,zhang2021transfuse}, pure-transformer encoders \cite{cao2021}, and gated axial attention \cite{valanarasu2021}. Boundary-aware transformers introduce contour information into skin lesion segmentation \cite{wang2021bat}. Related medical segmentation work uses statistical texture and complementary context \cite{xu2024skinformer,xu2021dcnet,xu2023drfl}, task-specific vessel and polyp representations \cite{wang2022danet,wang2021retinal,xu2024pstnet,wang2023polypcontext}, and graph-based spatial relationships \cite{xu2022biological,meng2021}. Thin vessels, individual instances, and a single lesion foreground require different annotation and boundary conventions.

\subsection{Supervision and Evaluation Conditions}
Medical segmentation has explored soft-mask supervision \cite{wang2022softgan,wang2023dsnet} and affinity-aware image-level supervision \cite{wang2024accam}. Weakly supervised semantic segmentation also uses activation maps, correspondence distillation, matting, and mutual representation learning \cite{xu2023wavecam,xu2023scd,wang2023matting,xu2025rml}; related object-localization studies address foreground discrimination and incomplete activation \cite{wang2024idc,xu2024tmt}. The distinction between a localization cue and a complete binary reference is consequential here: the mirror contains grayscale mask boundaries, so binarization forms part of the evaluation definition.

Segment Anything and its medical adaptation introduce promptable segmentation \cite{kirillov2023,ma2024}; other work extends segmentation to unseen classes and open vocabularies \cite{xu2024spectral,xu2024gba,yi2025ovbis}. Multimodal 3D segmentation and scene completion additionally change the available inputs and geometric target \cite{xu2024deffusion,xu2024mrf}. Comparisons must therefore specify supervision, adaptation data, prompt provenance, and output representation. Our image-only methods receive no reference-derived prompts.

\subsection{Computation and Pipeline Design}
nnU-Net treats preprocessing, configuration, training, and postprocessing as an integrated system \cite{isensee2021}. MALUNet and EGE-UNet target compact skin lesion networks \cite{ruan2022,ruan2023}; UNeXt combines convolution with tokenized MLP blocks for rapid medical segmentation \cite{valanarasu2022unext}. Efficient attention is a further direction in visual representation learning \cite{feng2026laplacian}. Parameter counts and theoretical operation counts do not directly determine CPU runtime, which also depends on implementation, resolution, and hardware. We measure computation within one fixed pipeline and retain the unrefined mask to establish whether refinement justifies its cost. Published network scores are not pooled with these measurements because the datasets and evaluation conditions differ.

\section{METHODS}
\label{sec:methods}
\subsection{Common Image Pipeline}
Each RGB image is resized to $128\times128$ pixels using Pillow's LANCZOS resampler and converted to float64 in $[0,1]$. Development compares blue intensity with luminance $0.2125R+0.7154G+0.0721B$. The selected channel is smoothed with a Gaussian of standard deviation one pixel and reflected boundaries, then supplied to every method. The pipeline uses no hair removal, border cropping, or image-specific parameter tuning.

Common postprocessing retains the largest 8-connected component and fills its enclosed holes. Nearest-neighbor interpolation returns the mask to the mirror's $248\times248$ grid for scoring. This procedure assumes a single dominant lesion.

\subsection{Threshold Baseline and Initializers}
Otsu's criterion maximizes between-class variance \cite{otsu1979}. We use 256 equal-width bins on $[0,1]$, select the first maximizing split in a tie, and threshold at the corresponding bin boundary. The darker class defines the lesion candidate. Applying common postprocessing to this mask gives the Otsu-only baseline.

We compare three initial level-set fields in Chan--Vese:
\begin{itemize}
\item \textbf{Checkerboard:} the upstream sinusoidal preset with five-pixel squares.
\item \textbf{Disk:} the upstream large-disk preset, centered at pixel $(63,63)$ with radius 63 on the processing grid.
\item \textbf{Otsu seed:} the unprocessed threshold mask $M$ is converted into
\begin{equation}
\phi_0(x)=\frac{d(x,M^c)-d(x,M)}{64},
\label{eq:init}
\end{equation}
where $d$ denotes Euclidean distance to the indicated set. Thus $\phi_0>0$ inside the threshold region. The denominator fixes the field scale relative to the image half-width.
\end{itemize}
The level-set scale in Equation~\ref{eq:init} is fixed throughout the study.

\subsection{Contour Evolution and Stopping}
For normalized image $I$ and contour $C$, the underlying two-region energy can be written as
\begin{align}
E(C,c_1,c_2)={}&\mu\operatorname{Length}(C)\nonumber\\
&+\lambda_1\int_{\mathrm{in}(C)}(I-c_1)^2\,dx\nonumber\\
&+\lambda_2\int_{\mathrm{out}(C)}(I-c_2)^2\,dx,
\label{eq:energy}
\end{align}
where $c_1,c_2$ are region means. The area term is absent. The implementation rescales its scalar input to $[0,1]$ using the image minimum and maximum. Its finite-difference updates, smoothed Heaviside/Dirac functions, boundary treatment, and evolution without level-set reinitialization are retained \cite{skimagecode}.

All contour runs use $\mu=0.25$, $\lambda_1=\lambda_2=1$, time step $0.5$, and a maximum of 500 iterations. The upstream stopping statistic is
\begin{equation}
r_t=\sqrt{|\Omega|^{-1}\sum_{x\in\Omega}(\phi_{t+1}(x)-\phi_t(x))^2}.
\label{eq:stop}
\end{equation}
Iterations stop when $r_t\leq\epsilon$ or the budget is exhausted. Primary runs use the upstream default $\epsilon=\epsd$. A one-step disk run observed during development motivated a sensitivity analysis at $\epsilon=\epst$ for all three initializers. This analysis was specified before test prediction and retained the selected channel and all other settings.

Because phase labels are arbitrary, the phase with lower mean processed intensity is designated as the lesion whenever both phases are nonempty. Empty or full masks are retained. Phase assignment and postprocessing use no reference-mask information.

\section{EXPERIMENTAL DESIGN}
\label{sec:experiment}
\subsection{Data, Splits, and Integrity}
The public \texttt{flyingU-ai/isic2017} repository, pinned to commit \texttt{4480fed25ccb}, contains 2,000 image-mask pairs in \texttt{train} and 600 in \texttt{val} \cite{mirror}. Development uses the 100 training paths with the smallest unsigned 32-bit FNV-1a hashes of \texttt{20260909:} concatenated with the path. All 600 validation pairs form the held-out set. The selection rule and case list were recorded before performance evaluation; image IDs do not overlap across sets.

Images and masks are paired by exact ISIC filename ID. All 1,400 PNG files match their expected Git blob SHA-1 and byte count and have dimensions $248\times248$. Every mask contains intermediate grayscale values and is binarized at $128$ on the 0--255 scale. The resulting references are nonempty and nonfull. All selected cases are retained.

Decoded RGB hashes identify two duplicate-image pairs within the test partition, leaving 598 distinct arrays. Their reference masks differ, with Dice agreement 0.956 and 0.905. No exact RGB duplicate crosses the development/test boundary. The primary analysis retains all 600 IDs; a sensitivity analysis averages scores within each duplicate group and weights the 598 distinct arrays equally. Patient identifiers are unavailable.

The mirror's transformation history and correspondence to the official test partition remain unverified. Evaluation therefore concerns this repository partition. The original challenge portal lists the 2017 segmentation data under CC0 \cite{isicdata}.

\subsection{Development and Frozen Evaluation}
One common channel is selected by mean image Dice, averaged equally over the four default-setting methods on development data. Blue scores 0.5720 versus 0.4958 for luminance and is fixed for all test experiments.

The primary contrast is mean Dice for Otsu-initialized minus checkerboard-initialized Chan--Vese at the default tolerance. Secondary contrasts use the Otsu-only and disk-initialized baselines. All tighter-tolerance configurations are reported as sensitivity results. The protocol, selected channel, and source SHA-256 hashes were recorded locally before test execution; this record is not an external preregistration.

\subsection{Measurements and Uncertainty}
Metric selection should reflect the target structure and intended use \cite{taha2015,maierhein2024}. Aggregation and implementation choices can also change the interpretation of segmentation scores \cite{reinke2024}. We therefore report overlap and boundary agreement together, with explicit averaging and tolerance conventions.

For predicted mask $P$ and reference $G$, the principal overlap measures are
\begin{equation}
\operatorname{Dice}=\frac{2|P\cap G|}{|P|+|G|},\qquad
\operatorname{IoU}=\frac{|P\cap G|}{|P\cup G|}.
\end{equation}
Scores are averaged per image. Additional measures are the proportion with Dice below 0.5 and boundary F1 within a two-pixel Euclidean tolerance on the mirror grid. Boundaries comprise mask pixels removed by one $3\times3$ erosion; precision and recall count pixels within the tolerance of the opposing boundary. The tolerance is defined in pixels, without a physical-distance interpretation.

Uncertainty is summarized by 95\% percentile intervals from 10,000 paired bootstrap resamples of image IDs, using seed 20260909. These intervals describe image-level sampling variability and do not account for unknown patient clusters, site shift, or channel-selection uncertainty. Secondary and sensitivity contrasts are descriptive, without multiplicity adjustment.

Prespecified strata use reference lesion area: below 10\%, 10--30\%, and above 30\% of image area. This information is used only for analysis. Qualitative examples are the image IDs nearest the 10th, 50th, and 90th percentiles of the default Otsu-seed minus checkerboard Dice difference, with ties broken by ID.

\subsection{Implementation and Timing}
The numerical source is \texttt{\_chan\_vese.py} from scikit-image v0.25.2, Git blob \texttt{31c3f1e72af4} \cite{skimagecode}. A compatibility module redirects one private dtype-utility import; a trailing newline is the only other source change. All inputs use float64. The original blob is verified, and the imported function agrees exactly with the preserved source for three synthetic initializers.

Experiments use an Intel Xeon Platinum 8370C CPU at 2.80 GHz with an eight-core quota, Python 3.12.14, NumPy 2.3.5, SciPy 1.17.0, and Pillow 12.3.0. Eight image workers each use one numerical-library thread. Process CPU time includes preprocessing, initialization, evolution, postprocessing, and output resizing; file loading and scoring are excluded. These measurements describe computation per case, not end-to-end latency. All 4,200 held-out prediction masks are retained, and their saved overlap statistics are independently verified by recounting pixels.

\section{RESULTS}
\label{sec:results}
\subsection{Overlap and Boundary Agreement}

Otsu alone gives the highest mean Dice (0.6897), IoU (0.6011), and boundary F1 (0.3079) among the tested configurations (Table~\ref{tab:all}). At the default tolerance, Otsu-initialized Chan--Vese exceeds checkerboard initialization by 0.0649 in mean Dice (95\% paired image-bootstrap interval [0.0452, 0.0860]), but falls below Otsu alone (difference $-0.0237$ [${-0.0314}$, ${-0.0156}$]). Contour refinement therefore reduces overlap relative to the threshold baseline.

\begin{figure}[tbp]\centering

\includegraphics[width=\columnwidth]{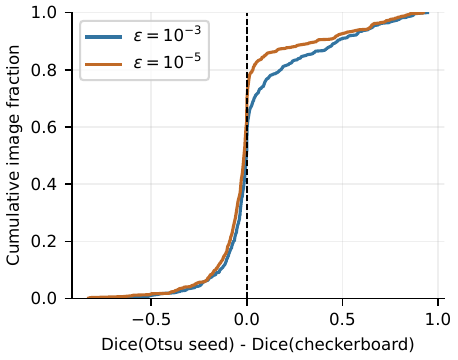}\caption{Cumulative distributions of per-image Dice differences. Positive values favor Otsu-seeded over checkerboard-initialized Chan--Vese.}\label{fig:cdf}

\end{figure}

The Otsu-seed versus checkerboard comparison improves on 257 images, ties on 30, and worsens on 313. Its positive mean difference reflects larger gains on a minority of images, as illustrated by the paired distributions in Figure~\ref{fig:cdf}.

\begin{table*}[!tb]\centering\small

\caption{Results for all 600 held-out image IDs. BF$_2$: boundary F1 within two pixels. Low Dice: count with Dice $<0.5$. Iteration count and CPU time are medians; timing excludes loading and scoring.}\label{tab:all}

\resizebox{\textwidth}{!}{%
\begin{tabular}{llccccrr}\toprule

Configuration & $\epsilon$ & Dice & IoU & BF$_2$ & Low Dice & Iter. & CPU (ms)\\\midrule

Otsu only & -- & 0.6897 & 0.6011 & 0.3079 & 135 & 0 & 3.8\\

CV: checkerboard & $10^{-3}$ & 0.6011 & 0.5049 & 0.1873 & 210 & 54 & 127.2\\

CV: disk & $10^{-3}$ & 0.3933 & 0.3113 & 0.0843 & 385 & 1 & 14.4\\

CV: Otsu seed & $10^{-3}$ & 0.6660 & 0.5654 & 0.2024 & 137 & 199 & 359.8\\

\midrule

CV: checkerboard & $10^{-5}$ & 0.6404 & 0.5491 & 0.2399 & 181 & 500 & 908.4\\

CV: disk & $10^{-5}$ & 0.4069 & 0.3314 & 0.1062 & 366 & 500 & 919.6\\

CV: Otsu seed & $10^{-5}$ & 0.6601 & 0.5590 & 0.1916 & 135 & 500 & 934.9\\

\bottomrule\end{tabular}}
\end{table*}

\begin{figure*}[!tb]\centering\includegraphics[width=\textwidth]{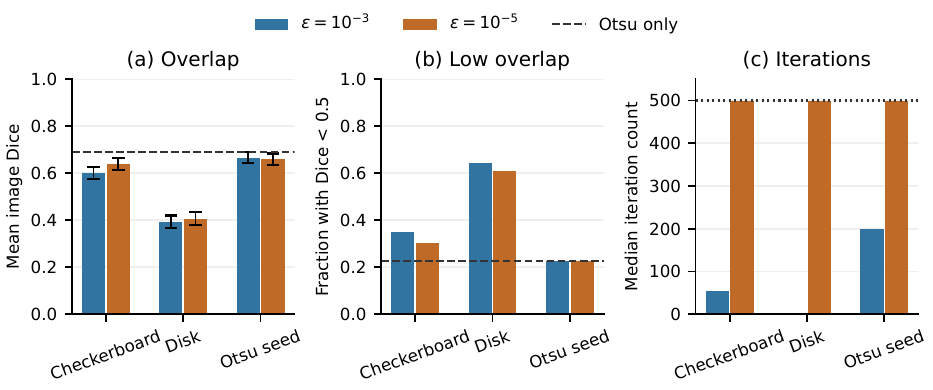}\caption{Overlap and computation under the two stopping tolerances. Dashed lines in (a,b) mark the Otsu-only baseline; error bars in (a) show 95\% image-bootstrap intervals. The dotted line in (c) marks the 500-iteration limit.}\label{fig:stopping}\end{figure*}

\begin{table}[tbp]\centering\scriptsize
\caption{Mean image Dice by reference foreground fraction. D: $\epsilon=10^{-3}$; T: $\epsilon=10^{-5}$.}\label{tab:size}

\resizebox{\columnwidth}{!}{%
\begin{tabular}{lrrr}\toprule Configuration & $<10\%$ & 10--30\% & $>30\%$\\\midrule

Images & 225 & 182 & 193\\\midrule

Otsu & 0.583 & 0.747 & 0.760\\

Checkerboard D & 0.299 & 0.757 & 0.806\\

Disk D & 0.080 & 0.363 & 0.787\\

Otsu seed D & 0.534 & 0.733 & 0.756\\

Checkerboard T & 0.402 & 0.772 & 0.795\\

Disk T & 0.073 & 0.404 & 0.798\\

Otsu seed T & 0.543 & 0.711 & 0.749\\

\bottomrule\end{tabular}}

\end{table}

\subsection{Stopping-Criterion Sensitivity}

At the default tolerance, disk initialization stops after one iteration in 471/600 cases (78.5\%). Median iteration counts are 54, 1, and 199 for checkerboard, disk, and Otsu initialization, respectively (Figure~\ref{fig:stopping}); 15, 7, and 87 runs reach the ceiling. Thus, the common maximum budget yields substantially different amounts of computation.

At $\epsilon=10^{-5}$, mean Dice is 0.6404 for checkerboard, 0.4069 for disk, and 0.6601 for Otsu initialization. All three have a median of 500 iterations. The limit is reached by 551 checkerboard runs and all 600 runs for each other initializer, leaving full convergence unresolved.

The Otsu-seed advantage over checkerboard decreases to 0.0197 [0.0006, 0.0396]. The change in this contrast is $-0.0453$ [${-0.0610}$, ${-0.0294}$]. Stopping tolerance therefore affects the observed initialization comparison, although the iteration ceiling prevents interpretation as a comparison of converged solutions. Otsu-seeded contours also remain below Otsu alone, with a mean difference of $-0.0296$ [${-0.0410}$, ${-0.0179}$].

Median CPU time increases from 127.2 to 908.4 ms for checkerboard and from 359.8 to 934.9 ms for Otsu initialization. Otsu alone takes 3.8 ms, retaining both an overlap and a computation advantage in this implementation.

\subsection{Size Strata, Duplicate Images, and Examples}

Lesion size changes the ordering of the initializers (Table~\ref{tab:size}). Below 10\% foreground area, default-tolerance Otsu initialization increases Dice from 0.299 to 0.534 relative to checkerboard. Above 30\%, the ordering reverses: 0.756 versus 0.806. These strata describe reference area rather than clinical subgroups.

Equal weighting of the 598 distinct RGB arrays gives default- and tighter-tolerance Otsu-seed versus checkerboard differences of 0.0652 and 0.0197. Accounting for the two exact-duplicate groups therefore leaves the observed ordering unchanged.

\begin{figure*}[!tb]\centering
\includegraphics[width=\textwidth]{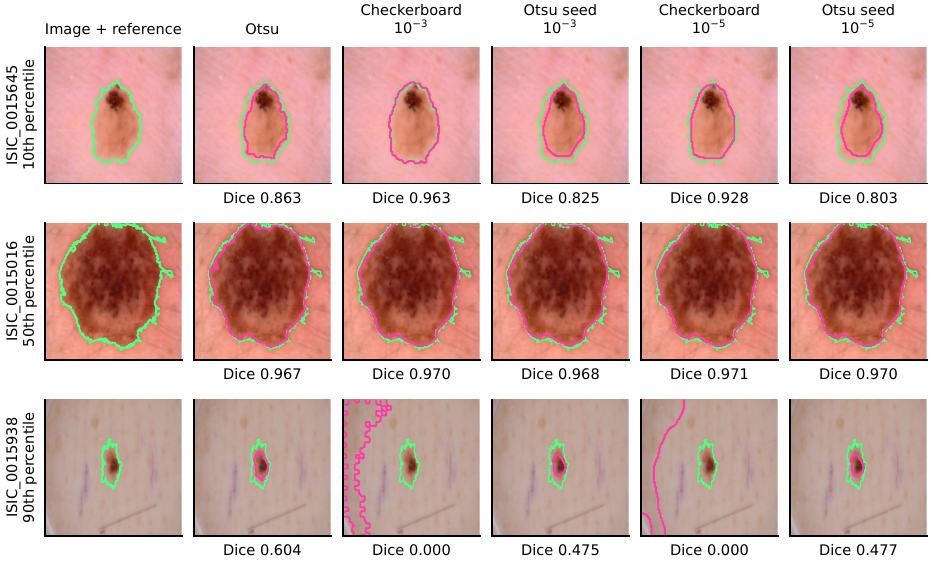}
\caption{Examples nearest the 10th, 50th, and 90th percentiles of the default Otsu-seed minus checkerboard Dice difference, shown from top to bottom. Green contours: reference; magenta contours: prediction. Values below panels are per-image Dice.}\label{fig:qual}
\end{figure*}

Figure~\ref{fig:qual} illustrates the fixed percentile-based examples. Checkerboard agrees better with the reference in the lower-difference example, while the median example gives similar masks. In the upper-difference example, checkerboard follows a peripheral intensity region; Otsu initialization localizes the lesion but underestimates its extent. A large improvement over checkerboard thus need not yield an accurate boundary.

\FloatBarrier

\section{DISCUSSION AND CONCLUSION}
\label{sec:discussion}
Otsu initialization improves mean Dice over checkerboard under both tolerances but reduces agreement relative to Otsu alone. The threshold comparator changes the interpretation of refinement, while lesion-size strata reveal a reversal hidden by the overall mean.

The one-step disk outcomes expose a limitation of reporting only the energy and maximum budget. Equation~\ref{eq:stop} measures field change, whose magnitude depends on geometry, scale, and update dynamics. A small update can coexist with poor segmentation. Tightening the tolerance changes the comparison, but ceiling hits leave convergence unresolved. Iteration distributions should accompany the tolerance and initial-field definition.

Shape and foreground scale also guide segmentation in remote sensing and infrared imagery \cite{xu2023rssformer,wang2023rsanet,xu2024hcf,xu2025samamba}; evaluating such priors for dermoscopy would require lesion-size strata alongside aggregate scores.

Extensions to embodied perception and action \cite{zhang2024navid,xu2025a0,zhang2026a1} would require temporal and task-success criteria beyond mask overlap.

The conclusions concern one global two-phase model, solver, curvature weight, and preprocessing pipeline. Local intensity models \cite{li2008rsf,lankton2008}, hair removal, trained networks, and adaptive initialization \cite{malik2022} require separate evaluation.

Resampling may alter texture and boundaries, and antialiased masks require explicit binarization. Missing patient, diagnosis, skin-tone, and acquisition metadata prevent patient-level independence checks and clinical subgroup assessment. Duplicate weighting addresses only the two detected RGB pairs. The darker-phase and largest-component assumptions can fail with border artifacts, disconnected regions, or low contrast. Boundary agreement alone does not establish clinical adequacy.

Thresholding achieves the strongest overlap at the lowest CPU cost in this cohort. Testing whether this ordering persists requires original-resolution data and independent datasets, retaining the unrefined baseline and recording numerical stopping behavior.

\begin{strip}
\section*{DECLARATIONS}
\noindent\textbf{Data and code.} This retrospective analysis uses publicly available image-mask files and involves no participant recruitment or intervention. The accompanying package contains the evaluated data, numerical source, configuration records, predictions, and per-image scores.

\noindent\textbf{Use of generative AI.} ChatGPT/Codex \cite{openai2026} assisted study design, literature retrieval, code development and execution, plotting, and manuscript preparation.
\end{strip}

\bibliographystyle{apalike-compact}
{\small\RaggedRight\interlinepenalty=10000\bibliography{references}}
\end{document}